\pdfoutput=1

\documentclass[journal]{IEEEtran}
\usepackage{tikz,multirow,amsmath,amssymb,pgfplots,subfigure}
\usetikzlibrary{matrix,chains,positioning,decorations.pathreplacing,arrows}

\usepackage{array}
\usepackage{hyperref}

\newcolumntype{P}[1]{>{\centering\arraybackslash}m{#1}}
\newcolumntype{L}[1]{>{\arraybackslash}m{#1}}

\newcommand{\RNum}[1]{\uppercase\expandafter{\romannumeral #1\relax}}

\ifCLASSINFOpdf
\else
\fi
\begin{document}

\title{Improved Deep Spectral Convolution Network For Hyperspectral Unmixing With Multinomial Mixture Kernel and Endmember Uncertainty}

\author{Savas~Ozkan,~\IEEEmembership{Member,~IEEE,}
        and~Gozde~Bozdagi~Akar,~\IEEEmembership{Senior Member,~IEEE}
\thanks{S. Ozkan and G.B. Akar are with the Department of Electrical and Electronics Engineering, Middle East Technical University, 06800, Ankara, Turkey.}}

\maketitle

\begin{abstract}

In this study, we propose a novel framework for hyperspectral unmixing by using an improved deep spectral convolution network (DSCN++)  combined with endmember uncertainty. DSCN++ is used to compute high-level representations which are further modeled with Multinomial Mixture Model to estimate abundance maps. In the reconstruction step, a new trainable uncertainty term based on a nonlinear neural network model  is introduced  to provide  robustness to endmember uncertainty. For the optimization of  the coefficients of the multinomial model and the uncertainty term, Wasserstein Generative Adversarial Network (WGAN) is exploited to improve stability and to capture uncertainty. Experiments are performed on both real and synthetic datasets. The results validate that the proposed method obtains state-of-the-art hyperspectral unmixing performance particularly on the real datasets compared to the
baseline techniques.

\end{abstract}


%
\IEEEpeerreviewmaketitle

\section{Introduction}

A hyperspectral imagery can provide comprehensive knowledge about the Earth surface in the form of narrow-band spectra that has been used in the variety of remote sensing applications. Indeed, the separation of constituent materials (i.e., endmembers) and their fractions (i.e., abundances) which contribute to the measured hyperspectral data is a fundamental step since the spectra of the materials can be subjected to complex interactions~\cite{yang2011blind,ozkan2017endnet}. As a result, they can be mixed in different fractions due to the limitations of hyperspectral sensors and aggravate the problem to be overcome under these scattering effects. Spectral unmixing aims to determine the fractions of the materials blindly from the mixture data that enables to analyze the data more easily~\cite{yang2011blind, heylen2011non, settle1996relationship, keshava2002spectral}. 

Even if the mixture data can be simplified with a linear formulation (i.e., linear mixing model)~\cite{dobigeon2016linear}, in real applications, the model remains weak and underestimates the solution, in particular for the cases such as atmospheric / lighting conditions and variability of the intrinsic material spectra. To straighten the model, nonlinear models reformulate the solution by employing nonlinearity and sparsity in their assumptions~\cite{dobigeon2016linear, heylen2016multilinear, somers2009nonlinear, altmann2011bilinear, nascimento2009nonlinear, raksuntorn2010nonlinear}. In the literature, the bilinear model is a commonly used technique for handling nonlinear scenarios where the bilinear interactions of linear abundances are taken into account in the model. As stated~\cite{dobigeon2016linear}, the methods based on the bilinear model mainly differ with the additional parameter terms and constraints exploited in their formulations.

However, both models assume that a hyperspectral scene is composed of a limited number of fixed endmember spectra. Hence, the performance of the methods for endmembers and their abundances may be limited because the possible spectral variability / endmember uncertainty resulting from the data is not taken into account. Fig.~\ref{fig:endvar} illustrates the spectra of pixels from the Jasper and Urban datasets~\cite{zhu2014structured, zhu2014spectral} that are labeled as pure materials. It can be seen that each material can form differently even for the same scene under the assumption of spectral variability. For this purpose, the conventional methods suffer from the disadvantages of the models (i.e., linear and non-linear models), and the representations are degraded by the use of false end members and their combinations. More recently, the angler-based loss function methods have achieved state-of-the-art performance that can partially overcome their limitations~\cite{ozkan2017endnet}. Similarly, the idea of stack autoencoders is adapted to learn blindly spectral signatures. This provides robustness for outliers and noise sensitive data due to the large set of trainable parameters~\cite{su2019daen}. Moreover, the methods~\cite{ozkan2017endnet, feng2018hyperspectral} explain the importance of sparsity for hyperspectral data. ~\cite{zhang2018hyperspectral} proposes a deep convolution network to estimate abundances while using both spectral and spatial neighborhoods. However, due to the lack of examples, the solution is overfitted.

In addition to the linear and nonlinear models, a variety of techniques consider the spectral variability in their formulations. In general, the definition of spectral variability assumption is simply a linear/non-linear combination of true endmembers  $\mathbf{e}_k \in \rm I\!R^{D}$ and abudances $y_k \in \rm I\!R^1 , k=1,...,K$ for each pixel $\mathbf{x} \in \rm I\!R^D$ so that there is a random uncertainty term $\mathbf{\gamma}_{k} \in \rm I\!R^{D}$ that also contributes to the material response~\cite{zhou2018gaussian}:
\begin{eqnarray}
\label{eqn:bc1}
\mathbf{x} = \sum_{k=1}^{K} {(\mathbf{e}_k + \mathbf{\gamma}_{k} ) y_k} + \eta, \hspace{2mm}  s.t. \hspace{1.2mm} y_k \geq 0,\hspace{1.2mm}  \sum_{k=1}^{K} {y_k} = 1
\end{eqnarray}

\noindent where $K$ is the number of endmembers exhibited in the scene and $D$ is the spectral bands of a pixel. Also, $\eta \sim \mathcal{N}(0,1)$  is the additive Gaussian noise to account the possible noise sources. Note that $\mathbf{\gamma}_{k}$ can vary for each pixel and material, regardless of the scene, which virtually defines the uncertainty of the data.

In the literature, the studies that intend to be robust to the endmember uncertainty are based on two different sets of assumptions. One of them exploits a library of on-hand material spectra (i.e., the set-based assumption) so as to determine abundances~\cite{roberts1998mapping, heylen2016hyperspectral, quintano2013multiple, roberts2003evaluation, dennison2003endmember} while the other uses the statistical distributions of the materials (i.e., the distribution-based assumption)~\cite{zhang2014pso, eches2010estimating, du2014spatial, zhou2018gaussian}. For the set-based assumption, the straightforward technique to represent the spectral variability is the multiple endmember spectral mixture analysis (MESMA)~\cite{roberts1998mapping}. This algorithm intuitively unmixes data iteratively by trying all possible material spectra presented in the library until the error/conditions are completely satisfied or reached to the desired values. The main drawback is that this method requires a lot of computational effort, especially when a large body of a spectra library is used. It is inevitable to conduct a brute-force search for every possible models is intractable. Its variants have an objective to reduce the computation complexity and makes the assumption more tractable. Most variants in the literature focus on reducing library size by selecting the representative spectra from the library~\cite{quintano2013multiple, roberts2003evaluation, dennison2003endmember}. The rest~\cite{heylen2016hyperspectral} modifies the search algorithm which incrementally determines the possible solutions from a subset of endmembers so that the complexity exhausted in the search step is moderately reduced. 

Similarly, the distribution-based models present the final endmembers as if they are from some distributions, so that the flexibility for the spectral variability can be improved. For this purpose, different distribution types can be used in the literature to represent the endmembers (i.e., normal composition model (NCM)~\cite{zhang2014pso} or beta composition model (BCM)~\cite{du2014spatial}). However, Zhou et. al.~\cite{zhou2018gaussian} states that the true representation of endmembers might not be approximated to an unimodal distribution and the mixture of sub-distributions (i.e., Gaussian Mixture Model (GMM)) may be more appropriate for modeling real data in particular. It should be noted, however, that the method needs high memory requirements which makes the algorithm impractical for any real scene, even at medium spatial resolutions. Furthermore, even if the expectation maximization is one of the prominent techniques for optimizing GMM coefficients~\cite{dempster1977maximum}, it suffers from convergence to global solutions and it is not reliable to learn the ideal multinomial distributions as explained in~\cite{dempster1977maximum, kolouri2017sliced, jian2011robust, amendola2015maximum, jin2016local}.

In this study, we extend the Deep Spectral Convolution network (DSCN) framework which is combined with the endmember uncertainty and the multinomial mixture kernel for hyperspectral unmixing. For this purpose, improved  DSCN (i.e., DSCN++) is exploited to compute high-level representations. Later, these representation are modeled with Multinomial Mixture Model to estimate abundance maps. In addition, a new trainable uncertainty term based on a nonlinear neural network model  is introduced in the reconstruction step to provide robustness to endmember uncertainty. For the optimization of  the parameters of the multinomial model and the uncertainty terms, Wasserstein Generative Adversarial Network (WGAN) is used to improve stability and to capture the uncertainty. Note that use of Wasserstein Generative Adversarial Network (WGAN) for the first time is fairly novel compared to baseline techniques due to its advantages and performance. To this end, hyperspectral data is unmixed and the abundances are estimated blindly from a set of precomputed endmember spectra in an end-to-end learning scheme.

In summary, the overall model takes hyperspectral data as input and computes low-dimensional representations by using Improved DSCN (DSCN++). Then, these representations are modeled with the Multinomial Mixture Model to estimate abundance maps. In the reconstruction step, abundance maps are fed to two NN modules along with the projection of endmember signatures. To this end, individual outputs are summed and the input is reconstructed.

The contributions presented in the manuscript can be summarized as follows: 
\begin{itemize}

\item First, we improve the Deep Spectral Convolution Network (DSCN) model~\cite{ozkan2018dscn} by introducing modifications to the architecture. This practically reduces the chance of vanishing gradient problem and enhances the selectivity of the filters. To this end, the trainable parameters tend to converge to a more stable solution which is one of the drawbacks highlighted in the baseline algorithm~\cite{ozkan2018dscn}.

\item Second, we introduce a multinomial mixture kernel based on a neural network (NN) architecture that estimates the abundances per-pixel by mimicking the mixture of Gaussian distributions. To improve the effectiveness, the latent features obtained from the DSCN are used as the input for the multinomial mixture kernel. In particular, the parameter set used for the kernel is optimized with the Wasserstein GAN~\cite{arjovsky2017wasserstein, gulrajani2017improved} for the first time which leads to more accurate and stable solutions than the expectation maximization, the Kullback-Leibler (KL)~\cite{hershey2007approximating} divergence or the mean square/absolute error (ME)~\cite{ledig2017photo}. Moreover, WGAN model mitigates the limitations of EM-like methods such as sensitivity for parameter initialization and discontinuity that reduces the performance of GMM models. To this end, a flexibility of the model is defined with  robustness to spectral variability.

\begin{figure}
\centering
\includegraphics[scale=0.4]{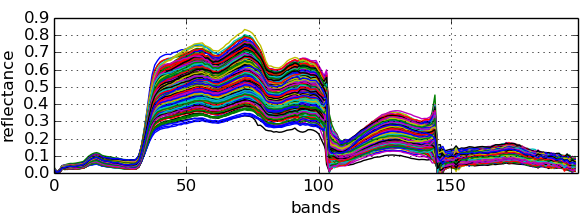}
\includegraphics[scale=0.4]{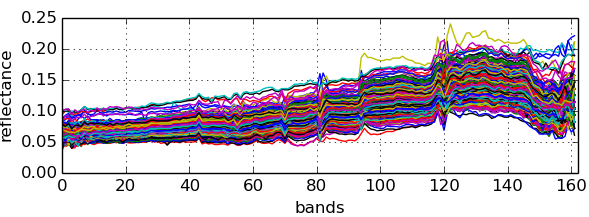}
\caption{Spectra of pixels from the Jasper (first row) and Urban (second row) datasets that correspondence to two different pure materials. It can be seen that hyperspectral data can be highly influenced by the assumption of spectral variability.}
\label{fig:endvar}
\end{figure}

\item Third, we define a new trainable uncertainty term based on a nonlinear NN model which simply provides robustness to the assumption of endmember uncertainty for the model. To do so, the estimated abundances are conditioned with an additional noise distribution as in~\cite{arjovsky2017wasserstein, gulrajani2017improved} to synthesize possible uncertainty. Furthermore, the similar architecture is exploited to minimize the residue estimated from precomputed endmember spectra in the reconstruction step. To this end, this auxiliary model practically helps to converge to a better solution for abundance estimation.

\item Note that all these modifications are formulated as a full hyperspectral unmixing pipeline with NN modules and it is optimized by a stochastic gradient-based solver in an end-to-end learning scheme. 

\end{itemize}

The rest of the manuscript is structured as follows. First, we define the concept of the DSCN and explain the modifications presented in the manuscript. Then, the details of the multinomial mixture kernel and endmember uncertainty are described. Later, the optimization step for learning trainable parameters are summarized. Lastly, the experimental results obtained on several real and synthetic datasets are presented and we conclude the manuscript.

\begin{figure}
\centering
\includegraphics[scale=0.26]{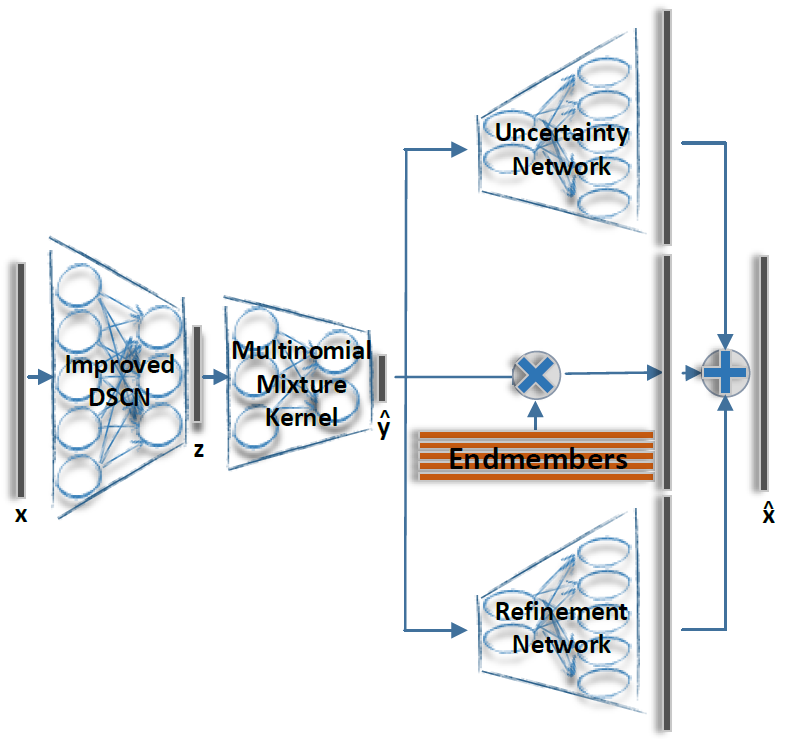}
\caption{Flow of the proposed method. This model takes hyperspectral data $\mathbf{x}$ as input and computes high-level representations $\mathbf{z}$ by using Improved DSCN. Then, these representations are modeled with Multinomial Mixture Model (i.e., optimized by WGAN) to estimate abundance maps $\mathbf{\hat y}$. In the reconstruction step, abundance maps are fed to two NN modules with the multiplication of endmember signatures. To this end, individual outputs are summed and the input is reconstructed.}
\label{fig:flow}
\end{figure}

\section{Improved Deep Spectral Convolution Network} \label{dscn}

In this section, we first provide a basic definition related to the Deep Spectral Convolution Networks (DSCNs). Then, the modifications for the architecture as well as the multinomial mixture kernel and the endmember uncertainty are explained in detail. Lastly, the optimization step is described. To ease the understandability, the flow of the proposed method is illustrated in Fig.~\ref{fig:flow}. 

The original DSCN architecture~\cite{ozkan2018dscn} is ultimately equivalent to the standard autoencoder formulation where it initially encodes the input spectra to a hidden latent feature $\mathbf{\hat y} = Enc(\mathbf{x}; \theta_e)$ and then decodes this latent feature to recompute the reconstructed version of the original spectra $\mathbf{\hat x} = Dec(\mathbf{\hat y}; \theta_d)$. Ultimately, iterative transformations are performed and the trainable parameter sets for both encoder $\theta_e$ and decoder $\theta_d$ parts are updated based on the reconstruction error calculated by a loss function(s). To this end, $\mathbf{\hat y}$ corresponds to the abundances per-pixel.

In particular, the main objective of the deep spectral convolution networks (DSCNs) is that on the contrary to linear layers~\cite{ozkan2017endnet}, it blindly extracts discriminative latent features over data by abstracting the local spectral characteristics. This is important because the dimensionality of the data limits the model to achieve an ideal solution through affine transformations (i.e., corresponds to matrix operations as explained in~\cite{ozkan2017endnet}) learned from unsupervised data. This can lead to irregularities / holes in the course of partitioning the feature space. To overcome this limitation, either supervised data~\cite{norouzi2011minimal, krizhevsky2012imagenet} or splitting the data into several overlapping/non-overlapping parts~\cite{jegou2011product} is needed to be utilized (the detailed information about this discussion can be found in~\cite{norouzi2011minimal, ozkan2018dscn}). Note that the assumption that the data is split into multiple overlapping / non-overlapping portions directly shares similar targets in the DSCNs with an unsupervised setting.

Furthermore, the DSCN architecture is mainly composed of 1D stacked convolutions (i.e., with normalization, non-linear activation and pooling layers) per-pixel to reveal the latent spectral characteristics of data. 

\begin{table}[b]
\begin{center}

\caption{Architecture of the improved DSCN. The notations of Conv1D(W, K, S) indicates 1D convolution
whose output dimension, kernel size and stride are W, K and S respectively. Furthermore, SN denotes Spectral Normalization, BN indicates Batch Normalization and AvgP(K) corresponds to Average Pooling layer with kernel size and stride of K. }

\begin{tabular}{c*{4}{c}}
\hline \hline
\textbf{Index} & \textbf{Index of Inputs} & \textbf{Operation(s)} & \textbf{Output Shape} \\
\hline
(1) & - & Input Spectra &  D \\
(2) & (1) & Conv1D(10, 21, 1) & D $\times$ 10 \\
(3) & (2) & LReLU + AvgP(5) + BN & D/5 $\times$ 10 \\
(4) & (3) & Conv1D(10, 3, 1) & D/5 $\times$ 10 \\
(5) & (3) & Conv1D(10, 5, 1) & D/5 $\times$ 10 \\
(6) & (3) & Conv1D(10, 7, 1) & D/5 $\times$ 10 \\
(7) & (4),(5),(6) & Concat & D/5 $\times$ 30 \\
(8) & (7) & LReLU + AvgP(2) + SN & D/10 $\times$ 30 \\
(9) & (8) & Conv1D(10, 3, 1) & D/10 $\times$ 10 \\
(10) & (9) & LReLU + AvgP(2) + SN & D/20 $\times$ 10 \\
(11) & (10) & Linear(M) + LReLU & M \\
\hline \hline
\end{tabular}
\label{tab:weight}

\end{center}
\end{table}

However, the lack of parameter convergence and parameter initialization are some issues in order to obtain stable results for the DSCNs~\cite{ozkan2018dscn}. Therefore, the architecture should be restructured so that the error calculated from the loss function is effectively propagated to the earlier layers in the model, otherwise the overfitting risk  for the architecture can be multiplied. Note that this problem is renowned as the vanishing gradient problem in the literature~\cite{krizhevsky2012imagenet, srivastava2014dropout}. To this end, the modified architecture for $\mathbf{z}=H(\mathbf{x}; \theta_{e})$ is illustrated in Table~\ref{tab:weight}.

For the clarity and understandability of the terminology, we divide the encoder part $Enc(\mathbf{x}; \theta_e)$ into two consecutive steps. First, the proposed DSCN module is applied to extract high-level latent features $\mathbf{z} = H(\mathbf{x}; \theta_{e})$ (i.e. DSCN++) where $\mathbf{z} \in \rm I\!R^{M}$ is the latent feature for each input spectra $\mathbf{x}$. Here, $M$ denotes the dimension of the latent feature and it can be tuned by the users (It is set to $10$ throughout the experiments). Then, these latent features $\mathbf{z}$ are used to estimate the abundances $\mathbf{\hat y} = G(\mathbf{z}; \theta_{e})$ (i.e. Multinomial Mixture Kernel with WGAN) by exploiting the proposed multinomial mixture kernel for the second step. 

For this purpose, initially, the parameter-free ReLU activation (called ReLU) is replaced by leaky rectified units (LReLU)~\cite{maas2013rectifier, he2015delving} in the model. This function is well studied in the literature, especially for regression problems, as this unit allows small yet non-zero gradients for negative responses of an input, even if it is not completely active. Moreover, the influence/resolution of negative responses is adjusted by a constant value determined as~\cite{maas2013rectifier}.

Second, the order of normalization layers (i.e. spectral normalization~\cite{ozkan2018dscn} or batch normalization~\cite{ioffe2015batch}) and spectral convolutions are reversed to reduce the chance of getting small/no activation(s). Note that spectral normalization rescales the input by merely leveraging  spectral statistics (i.e., mean and variance of spectral responses of data than batch characteristics). Therefore,  even if the normalized spectral values improve the selectivity of the representation particularly for the problem, it has also chance to increase the sparsity with the combination of ReLU~\cite{ozkan2018dscn} (i.e., the chance of vanishing gradient problem). In Table~\ref{tab:norm}, the percentage of active responses in the model by using the normalization layers before (ours) and after convolution operations on different datasets. It is clear that the proposed structure (i.e., pre-normalization) allows more information to forward to the next layers.

\begin{table}[b]
\begin{center}

\caption{Percentage of active responses in the model for Jasper and Urban datasets.}

\begin{tabular}{c*{3}{c}}
\hline \hline
Methods  & Jasper & Urban  \\
\hline

Post-Normalization &  42.93 & 39.18 \\
Pre-Normalization (ours)  & 51.79 & 48.42  \\

\hline \hline
\end{tabular}
\label{tab:norm}

\end{center}
\end{table}

In addition, we observe that after  initial convolution layer, applying batch normalization than spectral normalization leads to more accurate/stable representations since batch normalization can regularize the network as intended in Dropout layer~\cite{srivastava2014dropout} that reduces the overfitting of the trainable parameters. 

\begin{figure}[t]
\centering
\subfigure{
\label{fig:a}
\includegraphics[scale=0.17]{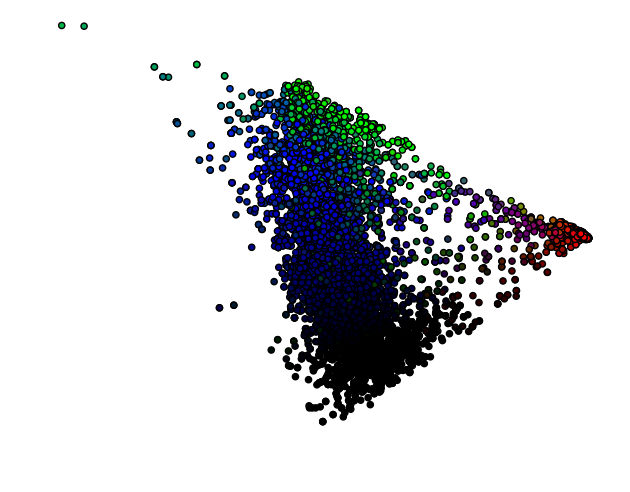}}
\subfigure{
\label{fig:b}
\includegraphics[scale=0.17]{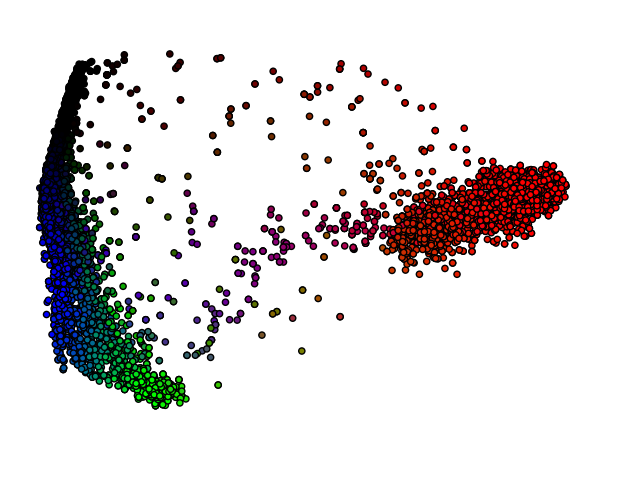}}
\subfigure{
\label{fig:a}
\includegraphics[scale=0.17]{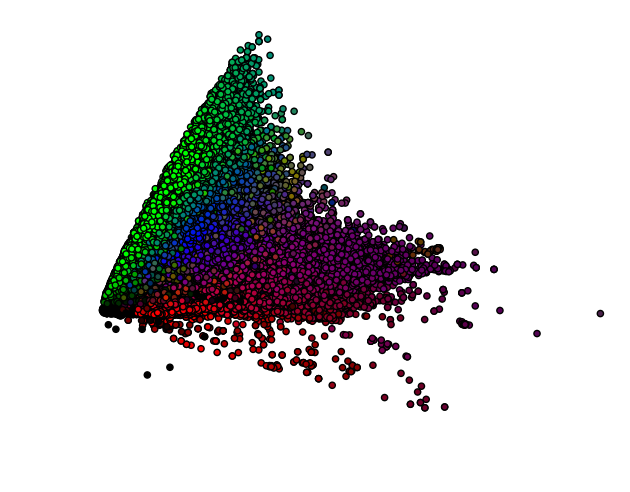}}
\subfigure{
\label{fig:b}
\includegraphics[scale=0.17]{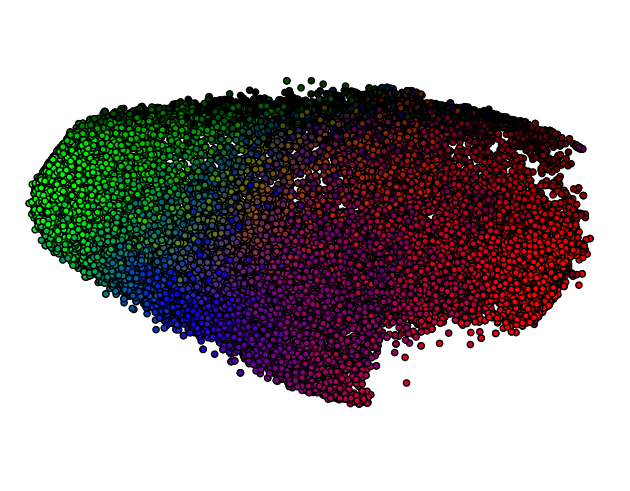}}
\subfigure{
\label{fig:a}
\includegraphics[scale=0.17]{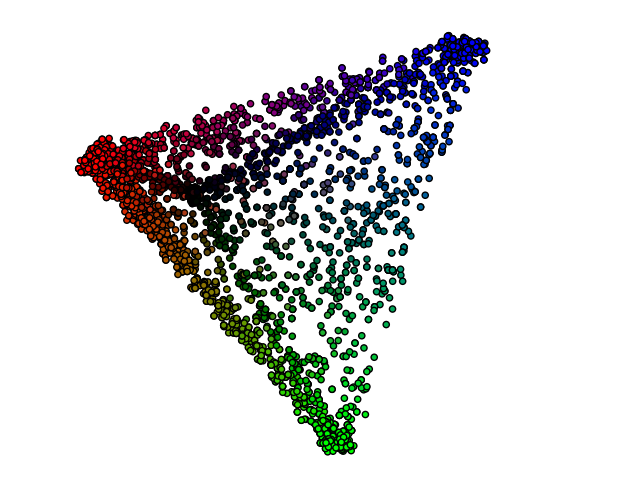}}
\subfigure{
\includegraphics[scale=0.17]{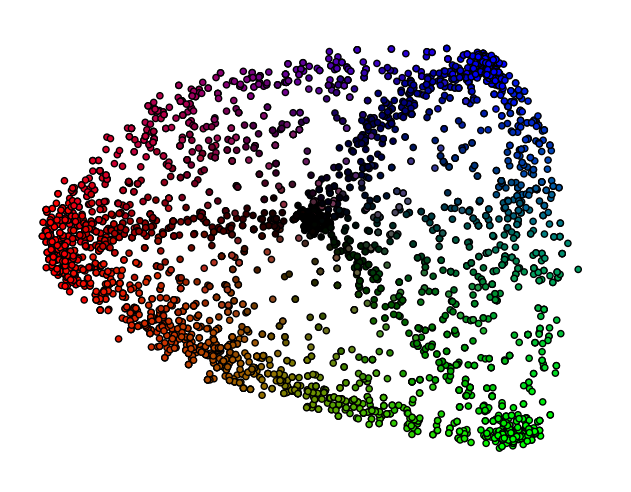}}

\caption{The data distributions before (first column) and after (second column) applying the improved DSCN to the input spectra for the Jasper (first row), Urban (second row) and Synthetic (third row) datasets. As seen, the latent feature space can be more separable than the original material spectra.}

\label{fig:subspace}
\end{figure}

When we consider the first convolution layer of the model as a low-level feature extractor, the consecutive layers aims to compute higher / distinct representations about data. Also note that the dimensionality of data various for different sensors. Inspired from~\cite{szegedy2016rethinking}, the inception module is adapted to enrich the capacity of the representation. In this way, multiple filters with different kernel sizes are performed for the same input and filters can select the best filters by reweighing the responses according to their importance for the data.

In addition, average pooling layers are leveraged to reduce the dimensionality of data at each layer. Note that the purpose of using an average operation instead of a max operation is to reduce the chance of overfitting to the constantly responding filters~\cite{srivastava2014dropout}. 

Note that WaveNet model applied for the synthesis of audio signals~\cite{oord2016wavenet} has a shared notion as in the proposed model and proves the success for 1D signals.

To demonstrate the effectiveness of the improved DSCNs, Figure~\ref{fig:subspace} visualizes the data distributions projected to 2D spaces before and after applying the improved DSCN. Note taht colors are discriminative based on their abundance values and the PCA is used to reduce the dimensionality.  It can be seen that the improved DSCN transforms the representation of data such that the latent features become more representative and the materials can be separable from the others. 

In the following section, we will explain the multinomial mixture kernel with WGAN loss that calculates the abundances per-pixel by exploiting  hidden latent features obtained by the improved DSCN.  

\subsection{Multinomial Mixture Kernel For Abundance Maps}

After high-level representations $ \mathbf{z}$ are computed from data $ \mathbf{x}$, corresponding abundances $ \mathbf{y}$  need to be obtained in an unsupervised manner. For this purpose, a NN-based multinomial mixture model is utilized. 

For the assumption of endmember uncertainty, compositional models sampled from an unimodal distribution estimate the distributions as follows: 
\begin{eqnarray}
\label{eqn:ncm}
p(y_k | \mathbf{z}) = \mathcal{N}(\mathbf{z} | \mathbf{\mu}_k,\,\mathbf{\Sigma}_k) ,
\end{eqnarray}

\noindent
Here, abundance maps $\mathbf{y}$ are estimated from given $\mathbf{z}$ by using a set of Gaussian distributions that are composed of mean $\mathbf{\mu}_k \in \rm I\!R^{M}$ and covariance matrix $\mathbf{\Sigma}_k \in \rm I\!R^{M \times M}$ where $k=1,...,K$.

As previously mentioned, these kernels can be misleading since the data cannot be directly approximated to an unimodal distribution~\cite{zhou2018gaussian} and multinomial mixture models need to be utilized to model the data properly.

For this purpose, Eq.~\ref{eqn:ncm} is generalized with Gaussian Mixture Model (GMM) by computing the distribution $p(y_k | \mathbf{z})$ from the statistical perspective:
\begin{eqnarray}
\label{eqn:gmm}
p(y_k | \mathbf{z}) = \sum_{n \in \mathcal{K}} p(n) p(y_k | \mathbf{z}, n),
\end{eqnarray}

\noindent 
Here, $n$ is a random variable whose sample space is $\mathcal{K}$. Moreover, the probabilistic values correspond to $p(n) = \pi_{k,n}$ and $p(y_k|\mathbf{z}, n) =  \mathcal{N}(\mathbf{z} | \mathbf{\mu}_{k,n},\,\mathbf{\Sigma}_{k,n}), n=1,...,N, k=1,...,K$. Hence, the model formulates the estimation of abundances such that each material abundance $y_k$ can be represented with the mixture of multivariate Gaussian distributions for a latent feature $\mathbf{z}$ as follows:
\begin{eqnarray}
\label{eqn:mg}
p(y_k | \mathbf{z})  = \sum_{n=1}^{N} \pi_{k,n} \mathcal{N}(\mathbf{z} | \mathbf{\mu}_{k,n},\,\mathbf{\Sigma}_{k,n})   \hspace{0.1mm}.
\end{eqnarray}

\noindent 
where $N$ is the number of components in the GMM and the constraints must be satisfied $\pi_{k,n} \geq 0,   \sum_{n=1}^{N} \pi_{k,n} = 1$. Moreover, $N \geq K$ assures that endmembers are overly represented with a mixture of latent Gaussian components that practically provides robustness for the endmember uncertainty. Moreover, $\pi_{k,n}$ is the weight term which determines the influence of the $n^{th}$ GMM component to the $k^{th}$ abundance value. 

Inspired by methods based on multinomial distributions, we claim that the distributions (i.e., components) can be imposed on abundances so that each abundance for a material can be overly expressed with the distributions. Note that this is ultimately related to the idea of spectral variability. For this purpose, novel NN kernel  and optimization function are proposed so that it computes the material abundances from the latent features $\mathbf{z}$ (i.e. Multinomial Mixture Kernel).

For our model, we mimic the GMM with a neural network (NN) module and a novel multinomial mixture kernel is proposed. This model is trainable in an end-to-end  manner and necessary parameters in the multinomial mixture kernel can be learned by error backpropagation. Note that a novel optimization function based on WGAN will be introduced in the subsection~\ref{wgan}

In particular, the distance between the latent representation $\mathbf{z}$ and $n^{th}$ distribution in Eq.~\ref{eqn:mg} can be  calculated simply by the multivariate mahalanobis distance:
\begin{eqnarray}
\label{eqn:umaha}
d_{k,n} = (\mathbf{\mu}_{k,n} - \mathbf{z})^T \mathbf{\Sigma}_{k,n}^{-1} (\mathbf{\mu}_{k,n} - \mathbf{z}).
\end{eqnarray}

By expanding the formulation in Eq.~\ref{eqn:umaha}, it can be roughly written as $d_{k,n} = \mathbf{z}^T\mathbf{\Sigma}^{-1}_{k,n} \mathbf{z} - 2 \mathbf{\mu}_{k,n}^T\mathbf{\Sigma}^{-1}_{k,n} \mathbf{z} + \mathbf{\mu}_{k,n}^T\mathbf{\Sigma}^{-1}_{k,n} \mathbf{\mu}_{k,n}$. Note that the formulation can be simplified with a linear neural network layer (i.e., with trainable weights $\mathbf{w}^1_{k,n}$ and bias $\mathbf{b}^1_{k,n}$) by losing the form of a quadratic function as $d_{k,n} = \mathbf{w}^1_{k,n} \mathbf{z} + \mathbf{b}^1_{k,n}$ where $\mathbf{w}^1_{k,n} \mathbf{z}= \mathbf{z}^T\mathbf{\Sigma}^{-1}_{k,n} \mathbf{z} - 2 \mathbf{\mu}_{k,n}^T\mathbf{\Sigma}^{-1}_{k,n} \mathbf{z}$ and $\mathbf{b}^1_{k,n} = \mathbf{\mu}_{k,n}^T\mathbf{\Sigma}^{-1}_{k,n} \mathbf{\mu}_{k,n}$. Furthermore, the distances for all endmembers is normalized with a sigmoid activation to produce a probabilistic relation (Please note that softmax distribution also works seamlessly/similarly for this model). To this end, the normal distribution is resembled with a cumulative distribution function as follows:
\begin{eqnarray}
\label{eqn:soft1}
 \mathcal{N}(\mathbf{z} | \mathbf{\mu}_{k,n},\,\mathbf{\sigma}_{k,n})  =    \frac{1}{1+e^{\mathbf{w}^1_{k,n} \mathbf{z} + \mathbf{b}^1_{k,n}}}.
\end{eqnarray}

Note that the kernel is implemented by directly using $\mathbf{\mu}_{k,n}$ and $\mathbf{\Sigma}_{k,n}$ coefficients (i.e. as a quadratic form) rather than parameterizing them with a linear NN layer. However, we observed that even if this model yields stable results (i.e., small variations), the overall accuracy is low. As explained in~\cite{arandjelovic2016netvlad}, this issue can be summarized by the fact that parameterizing the formulation with a linear NN layer makes the operations differentiable which enables to obtain better performance. However, the capacity of representing true relations (i.e., especially for the covariance matrix) is also underestimated. Therefore, parameterized coefficients are used through the manuscript.

Similarly, the weight term $\pi_{k,n}$ is calculated by a NN model and the sum-to-one constraint is supplied with a softmax activation:
\begin{eqnarray}
\label{eqn:soft2}
\pi_{k,n}  =  \frac{e^{\mathbf{w}^2_{k,n} \mathbf{z} + b^2_{k,n}}}{\sum_{n=1}^{N} e^{\mathbf{w}^2_{k,n} \mathbf{z} + b^2_{k,n}} }.
\end{eqnarray}

To this end, an abundance per-material is modeled as in Eq.~\ref{eqn:mg} by replacing the original components with the differentiable NN layers formulated with Eq.~\ref{eqn:soft1} and Eq.~\ref{eqn:soft2}. To assure the sum-to-one constraint for abundances of each pixel, L1-norm is applied at the end~\cite{ hapke1981bidirectional}.  

Last but not least, note that the proposed kernel decreases the drawback of NNs to noise, since multinomial mixture models can effectively capture the uncertainty in the data so that the trainable parameters yield more stable results as stated in~\cite{melchior2016filling, ozerov2012uncertainty}.

\subsection{Parameter Learning} \label{wgan}

To minimize the reconstruction error,  we exploit the optimization scheme proposed in~\cite{ozkan2017endnet}. For this purpose, the Spectral Angle Distance $C(\mathbf{x}, \mathbf{\hat x})$ between the input spectra $\mathbf{x}$ and the reconstructed spectra $\mathbf{\hat x}$ is maximized with the KL-divergence. Note that the similar notion (i.e., maximize the cross entropy) is used in GAN optimization:
\begin{eqnarray}
\label{eqn:bc5}
\begin{aligned}
\mathcal{L}_{re} =   -\mathbb{E}[KL\big( \hspace{0.1mm} 1.0 || C(\mathbf{x}, {\mathbf{\hat x}}) \hspace{0.1mm}  \big)] + \lambda_0 \mathbb{E}[|| \mathbf{x} - \mathbf{\hat x}||_1] \\
 + \lambda_1 \|  \mathbf{\hat y} \|_1 +  \lambda_2 \| \theta_e \|_2
\end{aligned}
\end{eqnarray}

\noindent 
where $\lambda_1$ determines the influence of the $l1$ regularization term (i.e., sparsity) and $\lambda_2$ is used for regularizing the parameters of the encoder module (except $\mathbf{w}^1$s, $\mathbf{b}^1$s) with $l2$ norm. In this work, $\lambda_1$ and $\lambda_2$ are set to $0.4$ and $10^{-5}$ respectively. Furthermore, $\lambda_0$ defines the ratio of the mean square/absolute error (ME) contributed to the reconstruction error. Note that ME is particularly critical for synthetic datasets since the angler distinctiveness is negligible compared to real data due to the additive noise (we will explain this statement in the experiment section in detail). Hence, in the default setting, $\lambda_0$ is set to 0. Lastly, $\mathbb{E}[.]$ denotes the expectation value of the possible values for the mini-batch.

In particular, the uncertainty exhibited from data cannot be captured with SAD and Euclidean-based loss functions. In the proposed method, it is modeled with WGAN loss by  providing theoretical and mathematical explanation.

First, the expectation maximization (EM) is the most popular technique to estimate GMM parameters in the literature and it theoretically guarantees to obtain a stationary solution by minimizing the negative log-likelihood function. However, the log-likelihood function highly suffers from to converge to a global solution by random parameter initialization and this assumption is highlighted in various studies in the literature. As explained in~\cite{dempster1977maximum, kolouri2017sliced, jian2011robust, amendola2015maximum, jin2016local}, this stems from the fact that solving negative log-likelihood function is similar to using the Kullback-Leibler (KL)-divergence and it has weakness that can be summarized as twofold. First, since GMM is continuous and locally Lipschitz~\cite{kolouri2017sliced}, the optimization method should also be continuous and differentiable, although this cannot be fully accomplished by the KL-divergence. Second, the KL-divergence suffers from the local minimum due to the discontinuity of the function which becomes sensitive to the choice of initial parameters. Further discussions related to the KL-divergence can be found in~\cite{kolouri2017sliced}.

\begin{table}[b]
\begin{center}

\caption{Architecture of Discriminative module. Differently, batch normalization and strides are used to obtain latent representations.}

\begin{tabular}{c*{4}{c}}
\hline \hline
\textbf{Index} & \textbf{Index of Inputs} & \textbf{Operation(s)} & \textbf{Output Shape} \\
\hline
(1) & - & Input Spectra &  D \\
(2) & (1) & Conv1D(5, 21, 5) & D/5 $\times$ 5 \\
(3) & (2) & BN + PReLU & D/5 $\times$ 5 \\
(4) & (3) & Conv1D(10, 5, 2) & D/10 $\times$ 10 \\
(5) & (4) & BN + PReLU & D/10 $\times$ 10 \\
(6) & (5) & Conv1D(20, 5, 2) & D/20 $\times$ 20 \\
(7) & (6) & BN + PReLU & D/20 $\times$ 20 \\
(8) & (7) & Linear(5) & D/20 $\times$ 5 \\
\hline \hline
\end{tabular}
\label{tab:disc}

\end{center}
\end{table}

Similarly, for various studies regardless of their applications, the shortcomings of ME for the regression problems  are also stated~\cite{ledig2017photo}. In short, the ME-based solutions yield overly smoothed results due to the averaging of all possible solutions in the manifold space so that the resultant outputs can be underestimated. Possible solutions for an inverse problem can be exponentially multiplied if only pixel-wise relations are used. Note that this assumption is built for the assessment of image quality~\cite{wang2003multiscale, wang2004image} and due to its weakness, ME cannot preserve the structural similarity of data. 

As stated, the loss function utilized in the proposed method is based on the KL-divergence and the ME by minimizing the reconstruction error. Inevitably, the drawbacks of the techniques as we previously discussed and uncertainty exhibited from data adversely affect the performance. For this purpose, we exploit the Wasserstein GAN~\cite{arjovsky2017wasserstein, gulrajani2017improved} to optimize the trainable parameters especially for the mixture model and the uncertainty terms.

Briefly, generative adversarial networks (GANs) have shown impressive performance for learning data distributions in various machine learning (ML) problems by their ability to transform a known distribution $p$ to an unknown data distribution $q$. However, the original GANs are potentially not continuous to minimize the error as explained in~\cite{arjovsky2017wasserstein} and the similar limitations can be observed as if the KL-divergence is used (i.e., discontinuity). Therefore, the Wasserstein GAN is exploited in the proposed model which has several theoretical benefits over the original GAN architecture~\cite{arjovsky2017wasserstein,  frogner2015learning, peyre2012wasserstein}. 

Formally, Wasserstein distance $W(p, q)$ is constructed to measure the distance between $p$ and $q$ (generating variables $\mathbf{x_p}$ and $\mathbf{x_q}$ respectively.) by using the Kantorovich-Rubinstein duality~\cite{villani2008optimal} and minimizes the distance between two distributions:
\begin{eqnarray}
\label{eqn:wasser}
\begin{aligned}
\substack{\\ max \\ \| f \|_L \leq 1} \bigg( \mathbb{E}_\mathbf{x_p}[f(\mathbf{x_p})] - \mathbb{E}_\mathbf{x_q}[f(\mathbf{x_q})] \bigg),
\end{aligned}
\end{eqnarray}

Note that $f(.)$ is a 1-Lipschitz function. Inspired by the structure of Variational AutoEncoder (VAE)~\cite{kingma2013auto}, this duality formulation can be adapted to our problem:
\begin{eqnarray}
\label{eqn:wass}
\begin{aligned}
\substack{\\ max \\ \| f \|_L \leq 1} \bigg( \mathbb{E}_\mathbf{x}[f(\mathbf{x})] - \mathbb{E}_{\mathbf{\hat x}}[\sum_{k=1}^{K}\mathbf{e}_k\sum_{n=1}^{N} \pi_{k,n} f(\mathbf{\hat x}_{k,n})] \bigg),
\end{aligned}
\end{eqnarray}

\noindent
where $\mathbf{e}_k$s and $\pi_{k,n}$ are not random variables. Moreover, $\mathbf{\hat x}_{k,n}$ corresponds to the generated sample from  $n^{th}$ GMM component for each material $k$. Thus, this formulation can be simplified as follows:
\begin{eqnarray}
\label{eqn:wasser}
\begin{aligned}
\substack{\\ max \\ \|f \|_L \leq 1} \bigg( \mathbb{E}_\mathbf{x}[f(\mathbf{x})] - \sum_{k=1}^{K}\mathbf{e}_k\sum_{n=1}^{N} \pi_{k,n} \mathbb{E}_{\mathbf{\hat x}_{k,n}}[f(\mathbf{\hat x}_{k,n})] \bigg),
\end{aligned}
\end{eqnarray}

In particular, the training process is stabilized by a gradient penalty~\cite{gulrajani2017improved} so that final loss function for Wasserstein Adversarial model is written as: 
\begin{eqnarray}
\label{eqn:soft}
\begin{aligned}
\mathcal{L}_{adv} = \mathbb{E}[D(\mathbf{x}; \theta_{adv})] - \mathbb{E}[D( \mathbf{\hat x};  \theta_{adv})] \hspace{1mm} \\
- \lambda_{pq} \mathbb{E}[(\| \triangledown_{\mathbf{\tilde x}} D(\mathbf{\tilde x};  \theta_{adv}) \|_2-1)^2].
\end{aligned}
\end{eqnarray}

\noindent 
where $D(.)$ is the discriminative module and it is composed of a family of 1-Lipschitz functions~\cite{kolouri2017sliced}. Moreover, $\mathbf{\tilde x}$ is sampled uniformly from $l1$ normalized input and reconstructed spectra (This is critical since the angler-based metric is used in the model). $\lambda_{pq}$ determines the influence of the gradient penalty~\cite{gulrajani2017improved} for stable parameter learning. 

Note that the Wasserstein distance is continuous and differentiable almost everywhere. Moreover, compared to the KL-divergence, the loss function is much smoother so that the method converges to the ideal solution freely from the initialization. Additionally, as discussed in~\cite{ledig2017photo}, more accurate results can be obtained, since it provides robustness to the averaging operations in the ME.

For the discriminative module $D(.)$, we adapt the terminology of the PatchGAN which is extensively used in the image translation problem~\cite{isola2017image}. More precisely, instead of penalizing the overall structure of spectra with the discriminative module, the spectral patches of spectra are used for the purpose of capturing local spectral characteristics of real and reconstructed spectra. The architecture details are presented in Table~\ref{tab:disc}.

\subsection{Spectral Variability Assumption} \label{eu}

In Eq.~\ref{eqn:bc1}, the formulation that encapsulates the spectral variability has, in comparison to the linear / nonlinear models, an additional term $\mathbf{\gamma}_k$ that defines the uncertainty per-pixel subjected to the abundances. In particular, this term should also be modeled with an additional parameter set in the model to obtain state-of-the-art performance, as it captures the possible deformations of spectra due to the spectral variability that cannot be accounted by the multinomial mixture kernel.

In the formulation, by simply expanding Eq.~\ref{eqn:bc1}, it is easy to see that there are two sets of equations where $\mathbf{e}_k \hat y_k$ denotes the projection of endmember spectra $\mathbf{e}_k$ (i.e., precomputed material spectra) with the estimated abundances $\hat y_k$ while the other term corresponds to $\mathbf{\gamma}_k \hat y_k$ that defines the influence of the uncertainty for each abundance. For this purpose, we introduce a trainable NN module $U(\mathbf{\hat y}, \eta; \theta_{u})$ which takes the estimated abundances $\mathbf{\hat y}$ and Gaussian noise $\eta \sim \mathcal{N}(0,1)$ as input and generates an additive nonlinear spectra. This spectra ultimately degrades the reconstructed spectra based on the uncertainty assumption. More specifically, the uncertainty module $U(.)$ consists of two consecutive nonlinear NN layers that incrementally increases the dimension of spectra (i.e., $K+L \rightarrow 20 \rightarrow D$) by exploiting the abundance values and Gaussian noise. Here, $L$ is the dimension of the noise. Final responses are rescaled with TanH activation as well as a trainable coefficient $\alpha_{u} \in [0, 0.1]$ to limit the influence. 

In addition, since the performance of hypersepctral unmixing is highly influenced by precomputed material spectra $\mathbf{\hat e}_k$ (i.e., it can be computed from various endmember extraction techniques), the representation of endmember spectra can be also improved with the model for better abundance results. This is reasonable because current state-of-the-art methods for endmember extraction are based on NN structure~\cite{ozkan2017endnet, zhang2018hyperspectral}. Instead of optimizing the precomputed material spectra by backpropagation, we have observed that introducing a new trainable term that refines the material spectra results in better performance.

For this purpose, the true material spectra is decomposed into $\mathbf{e}_k = \mathbf{\hat e}_k + \mathbf{r}_k$ where $\mathbf{\hat e}_k$ is the precomputed material spectra and $\mathbf{r}_k$ is the possible residue obtained by the endmember method. Similarly, this residue term can be modeled by exploiting the similar architecture as in the uncertainty with no shared parameters as $ \mathbf{r}_k = U(\mathbf{\hat y}; \theta_{r})$ and $\alpha_{r} \in [0, 0.05]$. Hence, Eq.~\ref{eqn:bc1} in the manuscript that corresponds to $\mathbf{\hat x} = Dec(\mathbf{\hat y}; \theta_d)$ can be rewritten as follows:
\begin{eqnarray}
\label{eqn:bc2}
\mathbf{\hat x} = \sum_{k=1}^{K} {\mathbf{e}_k \hat y_k } +  \alpha_{r} U(\mathbf{\hat y}; \theta_{r})
 +  \alpha_{u} U(\mathbf{\hat y}, \eta; \theta_{u})
\end{eqnarray}

In particular, parameters that capture the uncertainty of endmembers must be optimized with a novel scheme to achieve state-of-the-art performance. In the following section, we will introduce a novel loss function based on WGAN that is applied to hyperspectral unmixing problem for the first time.

\subsection{Implementation Details}

Since the proposed model consists of different modules, five individual parameter sets should be learned from data for parameter optimization. These are $\theta_g$ (corresponds to $\mathbf{w}_{k,n}^1$s, $\mathbf{b}_{k,n}^1$s), $\theta_e$, $\theta_r$, $\theta_u$ and $\theta_{adv}$. Therefore, loss functions and corresponding coefficient terms for each set are optimized as follows:
\begin{subequations}
\label{eqn:losses}
\begin{equation}
\mathcal{L}_{\theta_g} =    0.01 \hspace{0.4mm} \mathcal{L}_{re} + 0.1 \mathcal{L}_{adv},
\end{equation}
\begin{equation}
\mathcal{L}_{\theta_e} =    \mathcal{L}_{re},
\end{equation}
\begin{equation}
\mathcal{L}_{\theta_r} =    0.001 \hspace{0.4mm} \mathcal{L}_{re},
\end{equation}
\begin{equation}
\mathcal{L}_{\theta_u} =    0.001 \hspace{0.4mm} \mathcal{L}_{adv},
\end{equation}
\begin{equation}
\mathcal{L}_{\theta_{adv}} =    \mathcal{L}_{adv}.
\end{equation}
\end{subequations}

From Eq.~\ref{eqn:losses}, it can be seen that the Wasserstein GAN is used primarily to optimize the parameters of $\theta_g$ and $\theta_u$ that implicitly provides robustness to the parameters for spectral variability / uncertainty. GAN models can be powerful enough to generate a proper noise space based on the definition of the problem. This is critical since the uncertainty exhibited from hyperspectral data can be precisely captured by the Wasserstein distance. Moreover, Wasserstein distance has several theoretical benefits to obtain a stable and accurate solution as explained throughout the manuscript.

Moreover, to optimize the latent representations learned from data, the DSCN parameters are updated by exploiting only the reconstruction loss $\mathcal{L}_{re}$. In addition, the learning rates for multinomial mixing kernel and auxiliary functions are scaled with the constants to decrease the learning speed compared to the DSCN module. This is because DSCN parameters should converge faster to get more reliable abundances as expected.

Lastly, Adam stochastic optimizer~\cite{kingma2014adam} is utilized to update the parameters for each mini-batch by empirically defining the first-order moment term $\beta_1$ as 0.7. Note that the mini-batch size is fixed to 64. Also, we set the learning rate and the number of training iterations to 0.002 and 10K respectively for all datasets. The proposed algorithm is implemented on Python by extensively leveraging Tensorflow framework\footnote{Project Page: \url{https://github.com/savasozkan/dscn} }. Also, additional results are also provided.

\section{Experiments}
Experiments to compare the performance of the proposed method (DSCN++) with the baseline techniques are conducted on one synthetic and two real datasets. For this purpose, Linear Mixture Model (LMM)~\cite{heylen2016multilinear}, Generalized Bilinear Model (GBM)~\cite{halimi2011nonlinear}, Post-Nonlinear Mixing Model (PPNM)~\cite{altmann2012supervised}, Multilinear Mixing Model (MLM)~\cite{heylen2016multilinear} and Normal Compositional Model (NCM)~\cite{stein2003application} are selected for the hyperspectral unmixing baselines that are extensively used methods in the literature (codes are available on the web.). Note that the recent method in~\cite{zhou2018gaussian} needs high memory requirement which is impossible to compute the algorithm on real datasets especially by maintaining their original spatial resolutions.

Furthermore, Spatial Compositional Model (SCM)~\cite{zhou2016spatial}, Distance-MaxD (DMaxD)~\cite{heylen2011non} and Sparse AutoEncoder Network (EndNet)~\cite{ozkan2017endnet} are exploited to estimate endmember spectra from the scenes.

To evaluate the unmixing performance and compare the estimated abundances with the ground truth, Root Mean Square Error (RMSE) is utilized:
\begin{eqnarray}
\label{eqn:bc4}
RMSE(\mathbf{y}, \mathbf{\hat y}) = \sqrt{\frac{1}{L} \| \mathbf{y} - \mathbf{\hat y} \|^2_2}.
\end{eqnarray}

\noindent 
where $\mathbf{y}$ denotes the true abundance per-pixel while $\mathbf{\hat y}$ indicates the estimated ones by the different methods. Also, $L$ corresponds to the total number of pixels in the scene.

\begin{figure}[t]
\centering
\subfigure{
\label{fig:a}
\includegraphics[scale=0.4]{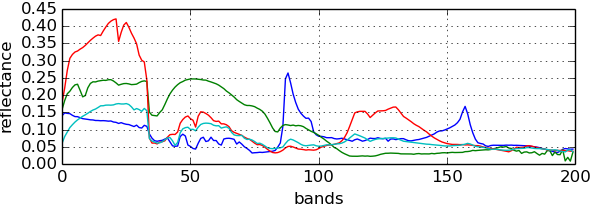}}

\subfigure{
\label{fig:a}
\includegraphics[scale=0.9]{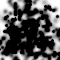}}
\subfigure{
\label{fig:b}
\includegraphics[scale=0.9]{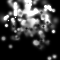}}
\subfigure{
\label{fig:a}
\includegraphics[scale=0.9]{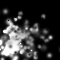}}
\subfigure{
\label{fig:b}
\includegraphics[scale=0.9]{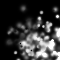}}

\caption{Endmember spectra (first row) and abundance maps (second row) for the Synthetic dataset. }

\label{fig:syntabun}
\end{figure}

In following sections, we first provide the quantitative performance for each dataset that clearly demonstrates the impact of the proposed method by highlighting each individual step as well as different parameter configurations in detail. Note that $N$ is the only coefficient needed to be tuned by the user. Therefore $N$ is determined for each step based on their overall results in the experiments (As expected, $N$ (the number of mixture components) can vary for each dataset as well as each endmember method based on its distinctiveness). Later, the comparisons with the baselines are explained. For reliability of assessments, tests are repeated 20 times for the proposed method, thus mean and standard deviation are reported.

To ease the understandability, some abbreviations used in the Tables can be summarized as follows: DSCN indicates the deep spectral convolution network as explained in the section \ref{dscn}.  WGAN indicates the use of Wasserstein GAN in the parameter optimization as mentioned in the subsection \ref{wgan}. Lastly, EU stands for the usage of NN models for the uncertainty term and the auxiliary model described in the subsection \ref{eu}.

\begin{table}[b]
\begin{center}

\caption{RMSE performance on the synthetic dataset for different configurations.}

\begin{tabular}{c*{5}{c}}
\hline \hline
$\times 10^{-2}$ & DMaxD~\cite{heylen2011non} & SCM~\cite{zhou2016spatial} & EndNet~\cite{ozkan2017endnet} \\
\hline

$N=4$   & 8.32 $\pm$1.4 & 4.47 $\pm$0.7 & 9.20 $\pm$1.4 \\
$N=8$   & 8.15 $\pm$1.4 & 4.36 $\pm$0.9 & 9.14 $\pm$1.0 \\
$N=16$ & 8.18 $\pm$1.0 & 3.97 $\pm$0.5 & 9.06 $\pm$1.1 \\
$N=24$ & 7.55 $\pm$1.1 & 4.42 $\pm$0.6 & 9.51 $\pm$1.2 \\
\hline
\hline
w$\setminus$o DSCN  & 10.78 $\pm$1.2 & 10.96 $\pm$1.5 & 13.37 $\pm$1.9 \\
w$\setminus$o EU       & 5.43  $\pm$0.3  & 3.81 $\pm$0.9  & 6.43 $\pm$0.4 \\
w$\setminus$o WGAN & 5.23  $ \pm$0.4 & 3.22 $\pm$1.2  & 8.88 $\pm$0.3  \\

\hline \hline
\end{tabular}
\label{tab:synt_conf}

\end{center}
\end{table}

\subsection{Synthetic Dataset}

\begin{figure}
\centering
\includegraphics[scale=0.4]{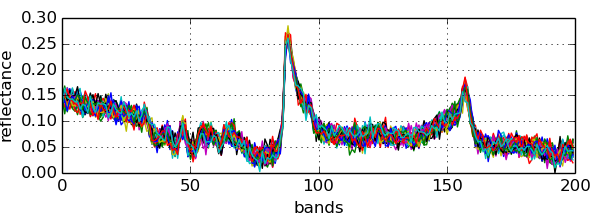}
\includegraphics[scale=0.4]{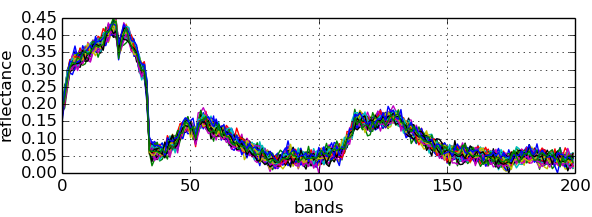}
\caption{Spectra of pixels from the Synthetic dataset. It can be seen that even if random noise is added in the construction of data, spectral variability cannot be exhibited truly and it can be misleading for learning-based methods since learning-based methods (generative-based models in particular) can learn this random noise easily.}
\label{fig:syntadverse}
\end{figure}

The unmixing methods are tested on the synthetic dataset released for the experiments in~\cite{zhou2018gaussian}. The dataset consists of four constituent materials from the ASTER spectral library as limestone, basalt, concrete and asphalt. The spectral range of data is from $0.4\mu$m to $14 \mu$m and the spectral dimension is resampled to 200. Furthermore, the spatial resolution of the scene is $60 \times 60$. For the scene, limestone is set as the background material, the rest is randomly placed to the scene by maintaining the Gaussian-shaped distributions. The endmember spectra and abundance maps for the scene are illustrated in Fig.~\ref{fig:syntabun}.

Note that even if synthetic datasets are conventional to test the hyperspectral unmixing performance, it is not quite possible to capture all adverse conditions observed in the real data (such as spectral variability). Hence, it can be hard to validate the performance truly under this strong bias. To degrade the bias, the only practical solution is to add random noise on the data which cannot reflect the true physical conditions (i.e., spectral variability) observed in the real data. More precisely, Fig.~\ref{fig:endvar} and Fig.~\ref{fig:syntadverse} show the severe difference between real and synthetic datasets for pure materials. It can be seen that the angler distinctiveness is quite small (i.e., spectra of pixels are nearly concentrated on a single response) and use of angler-based metric can be misleading for the dataset. Hence, only for the synthetic dataset, the KL-divergence with SAD in the reconstruction loss is replaced (for SCM) or jointly used (for DmaxD, Endnet) with the ME (i.e., $\lambda_0=1$).

The experimental results for each individual step and different parameter configuration settings are reported in Table~\ref{tab:synt_conf}. From the results, the proposed method achieves significantly better performance without EU and WGAN steps. This is quite reasonable since the data is degraded by random noise to simulate the spectral variability and the proposed model ultimately learns the noise from data than the assumption of spectral variability. Hence, this leads to the corruption for the model. However, note that true impacts of these steps can be clearly observed for real datasets. Therefore, further discussions about these steps are reserved for the real datasets.

The RMSE performance for the baselines are shown in Table~\ref{tab:synt_base}. From the results, the proposed method obtain slightly better performance compared to the baselines. Note that the proposed method is effective on the synthetic dataset even if it does not exploit all its features (i.e., EU and WGAN).

Lastly, note that ~\cite{zhou2018gaussian} obtains $1.53 \times 10^{-2}$ RMSE performance for this synthetic data. However, due to its computation requirement and differences for the endmember estimation, it is not fully compatible with our method.

\begin{table}[t]
\begin{center}

\caption{RMSE performance on the synthetic dataset for baselines. The best results are indicated as Bold-Red while the second best results are shown as Bold.}

\begin{tabular}{c*{4}{c}}
\hline \hline
$\times 10^{-2}$ & DMaxD~\cite{heylen2011non} & SCM~\cite{zhou2016spatial} & EndNet~\cite{ozkan2017endnet} \\
\hline

LMM~\cite{heylen2016multilinear} & \textbf{5.17} & 3.36 & 17.62 \\
GBM~\cite{halimi2011nonlinear} & 5.62 & \textbf{3.26} & 17.39 \\
PPNM~\cite{altmann2012supervised} & 7.66 & 4.01 & 9.34\\
MLM~\cite{heylen2016multilinear} & 6.52 & 3.85 & \textbf{6.71} \\
NCM~\cite{stein2003application} & \textbf{\textcolor{red}{4.09}} & 3.85 & 11.17 \\
\hline
DSCN++ & 5.23 & \textbf{\textcolor{red}{3.22}} & \textbf{\textcolor{red}{6.43}}  \\

\hline \hline
\end{tabular}
\label{tab:synt_base}

\end{center}
\end{table}

\begin{figure}[b]
\centering
\subfigure{
\label{fig:a}
\includegraphics[scale=0.4]{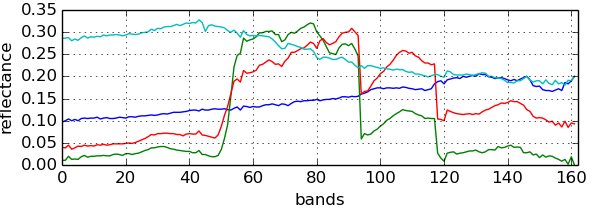}}

\subfigure{
\label{fig:a}
\includegraphics[scale=0.18]{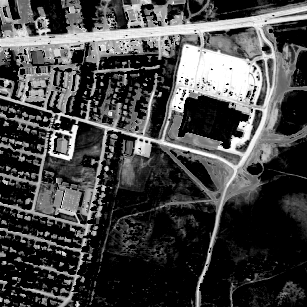}}
\subfigure{
\label{fig:b}
\includegraphics[scale=0.18]{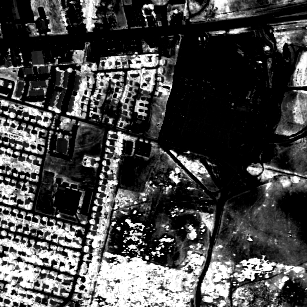}}
\subfigure{
\label{fig:a}
\includegraphics[scale=0.18]{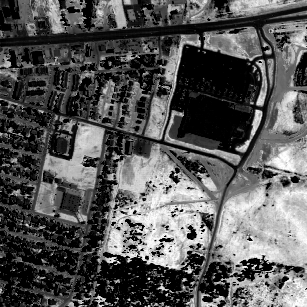}}
\subfigure{
\label{fig:b}
\includegraphics[scale=0.18]{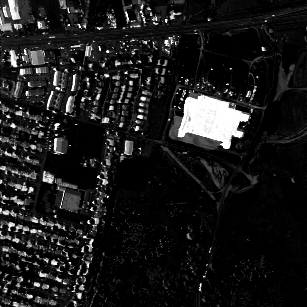}}

\caption{Endmember spectra (first row) and abundance maps (second row) for the Urban dataset.}

\label{fig:urbanabun}
\end{figure}

\subsection{Urban Dataset}

The spatial resolution of the scene is $307 \times 307$. The spectral range covers from 10 nm to 2500 nm and it is composed of 210 spectral bands. Several spectral bands (1-4, 76, 87, 101-111, 136-153 and 198-210) are removed from data due to the water-vapor absorption and atmospheric effects, thus the final spectral band is reduced to 162. Four material types are observed in the scene: asphalt, grass, tree and roof~\cite{zhu2014spectral}.  The endmember spectra and abundance maps for the scene are illustrated in Fig.~\ref{fig:urbanabun}.

The performance  for the proposed method under different configurations is illustrated in Table~\ref{tab:urban_conf}. It can be seen that the selection of $N$ larger than $K$ improves the performance for all endmember techniques. Note that the optimum $N$ value can vary since the discriminative power of the representation can be highly influenced by the endmember extraction techniques regardless of how spectral variability affects the scene. In addition, the results confirms that use of DSCN in order to reduce the dimension of data and to obtain compact representations yields better results. Moreover, for the models without the DSCN, the variations in the performance are significantly increased due to the fact that high dimensionality can lead to irregularities/holes for the space as previously described.

\begin{table}[t]
\begin{center}

\caption{RMSE performance on the Urban dataset for different configurations.}

\begin{tabular}{c*{4}{c}}
\hline \hline
$\times 10^{-2}$ & DMaxD~\cite{heylen2011non} & SCM~\cite{zhou2016spatial} & EndNet~\cite{ozkan2017endnet} \\
\hline

$N=4$ & 15.96 $\pm$2.6 & 13.26 $\pm$0.6 & 8.27 $\pm$0.4 \\
$N=8$ & 15.47 $\pm$1.9 & 13.29 $\pm$0.8 & 8.05 $\pm$0.5 \\
$N=16$ & 16.32 $\pm$3.2 & 13.24 $\pm$0.5 & 8.16 $\pm$0.4 \\
$N=24$ & 16.56 $\pm$2.1 & 13.23 $\pm$0.8 & 8.04 $\pm$0.3 \\
\hline \hline

w$\setminus$o DSCN  &  22.12 $\pm$2.6   & 16.63 $\pm$4.6    & 11.42 $\pm$0.8 \\
w$\setminus$o EU &  34.14 $\pm$0.3  & 15.91 $\pm$0.3    & 8.38 $\pm$0.4  \\
w$\setminus$o WGAN       &  31.43 $\pm$1.6   & 15.76 $\pm$1.0  & 8.52 $\pm$0.5 \\
\hline \hline
\end{tabular}
\label{tab:urban_conf}

\end{center}
\end{table}

\begin{table}[t]
\begin{center}

\caption{RMSE performance on the Urban dataset for baselines. The best results are indicated as Bold-Red while the second best results are shown as Bold. }

\begin{tabular}{c*{4}{c}}
\hline \hline
$\times 10^{-2}$  & DMaxD~\cite{heylen2011non} & SCM~\cite{zhou2016spatial} & EndNet~\cite{ozkan2017endnet} \\
\hline

LMM~\cite{heylen2016multilinear} & 31.04 & 28.00 & 27.59 \\
GBM~\cite{halimi2011nonlinear} & 31.04 & 27.00 & 27.08 \\
PPNM~\cite{altmann2012supervised} & 36.54 & 28.17 & 15.06\\
MLM~\cite{heylen2016multilinear} &  31.43 & \textbf{16.34} & \textbf{9.48} \\
NCM~\cite{stein2003application} & \textbf{31.04} &  28.00 & 27.51 \\
\hline
DSCN++ & \textbf{\textcolor{red}{15.47}} & \textbf{\textcolor{red}{13.23}} & \textbf{\textcolor{red}{8.04}} \\

\hline \hline
\end{tabular}
\label{tab:urban_base}

\end{center}
\end{table}

Similarly, considering endmember uncertainty with additional NN models practically improves the performance. For the Wasserstein GAN, it helps to converge to the ideal solution more likely. Moreover, it achieves more stable results by which the influence of parameter initialization is decreased in the model.   

The performance is compared with the baselines in Table~\ref{tab:urban_base} and the proposed method outperforms all hyperspectral unmixing techniques significantly. In particular, the usage of WGAN and EU steps improve the performance of DmaxD such that it leads to a highly accurate solution where the baselines stuck to stationary results. 

\begin{figure}[b]
\centering
\subfigure{
\label{fig:a}
\includegraphics[scale=0.4]{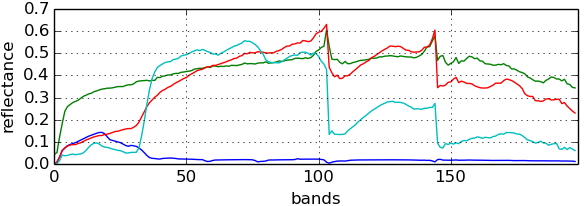}}

\subfigure{
\label{fig:a}
\includegraphics[scale=0.5]{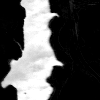}}
\subfigure{
\label{fig:b}
\includegraphics[scale=0.5]{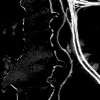}}
\subfigure{
\label{fig:a}
\includegraphics[scale=0.5]{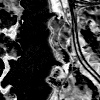}}
\subfigure{
\label{fig:b}
\includegraphics[scale=0.5]{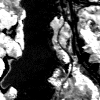}}

\caption{Endmember spectra (first row) and abundance maps (second row) for the Jasper dataset.}

\label{fig:jasperabun}
\end{figure}

\subsection{Jasper Dataset}

The Jasper dataset is captured by the AVIRIS sensor and the dimension is $100 \times 100$ pixels. Several channels (1-3,  108-112,  154-166 and 220-224) are discarded due to the atmospheric effects and water-vapor absorption. The spectral band is of 198. There are four constituent materials in the scene as tree, water, soil and road~\cite{zhu2014structured}. Fig.~\ref{fig:urbanabun} shows the endmember spectra and abundance maps for the scene. 

The detailed performance for different configuration settings and steps are reported in Table~\ref{tab:jasper_conf}. It can be seen that similar results can be observed as stated for the Urban dataset. A single outlier is that SCM without the DSCN achieves slightly better performance compared to the baseline. This stems from the fact that the SCM representation space can be more discriminative than the DSCN for this dataset. 

\begin{table}[b]
\begin{center}

\caption{RMSE performance on the Jasper dataset for different configurations.}

\begin{tabular}{c*{4}{c}}
\hline \hline
$\times 10^{-2}$ & DMaxD~\cite{heylen2011non} & SCM~\cite{zhou2016spatial} & EndNet~\cite{ozkan2017endnet} \\
\hline

$N=4$   & 12.03 $\pm$0.7 & 18.10 $\pm$0.9 & 6.38 $\pm$0.7 \\
$N=8$   & 12.29 $\pm$0.9 & 18.33 $\pm$1.1 & 6.61 $\pm$0.7 \\
$N=16$ & 12.12 $\pm$0.7 & 17.92 $\pm$0.8 & 6.31 $\pm$0.7 \\
$N=24$ & 12.27 $\pm$0.6 & 18.03 $\pm$1.1 & 6.62 $\pm$1.1 \\
\hline \hline
w$\setminus$o DSCN  &   12.47 $\pm$0.6  & 17.49 $\pm$1.2 & 6.87 $\pm$2.2 \\
w$\setminus$o EU &  14.78  $\pm$0.4  & 20.26 $\pm$0.4 & 8.36 $\pm$0.5  \\
w$\setminus$o WGAN       &  15.63 $\pm$0.3   & 19.03 $\pm$0.4 & 10.15 $\pm$0.5 \\
\hline \hline
\end{tabular}
\label{tab:jasper_conf}

\end{center}
\end{table}

\begin{table}[b]
\begin{center}

\caption{RMSE performance on the Jasper dataset for baselines. The best results are indicated as Bold-Red while the second best results are shown as Bold.}

\begin{tabular}{c*{4}{c}}
\hline \hline
$\times 10^{-2}$ & DMaxD~\cite{heylen2011non} & SCM~\cite{zhou2016spatial} & EndNet~\cite{ozkan2017endnet} \\
\hline

LMM~\cite{heylen2016multilinear} & 15.31 & 19.78 & 22.60 \\
GBM~\cite{halimi2011nonlinear} & 15.98 & 19.95 & 21.91\\
PPNM~\cite{altmann2012supervised} & \textbf{12.99} & \textbf{19.78} & 14.72 \\
MLM~\cite{heylen2016multilinear} &  16.59 & 20.73 & \textbf{8.13} \\
NCM~\cite{stein2003application} & 15.29 &  19.56 & 22.42 \\
\hline
DSCN++ & \textbf{\textcolor{red}{12.03}} & \textbf{\textcolor{red}{17.92}} & \textbf{\textcolor{red}{6.31}} \\

\hline \hline
\end{tabular}
\label{tab:jasper_base}

\end{center}
\end{table}

For the baselines comparisons, the proposed method obtains state-of-the-art performance as demonstrated in Tab~\ref{tab:jasper_base}. Compared to~\cite{ozkan2018dscn} which corresponds to the baseline DSCN, the improved model significantly improves the performance while small variations in the performance can be observed.

\section{Discussion and Conclusion}

In this manuscript, we introduce a novel neural network (NN) framework for hyperspectral unmixing. It is mainly composed of deep spectral convolution networks (DSCNs) by addressing the assumption of spectral variability that is one of the severe physical conditions frequently observed in real data. More precisely, in real applications, pixel responses for each material can be varied due to the atmospheric/illumination conditions instead of fixed material spectra. Therefore, these conditions should be addressed by an effective model in order to obtain state-of-the-art unmixing performance.

For this purpose, we present critical contributions throughout the manuscript as follows: First, we make several modifications in the architecture of the DSCNs to obtain stationary yet more accurate results. Second, a novel NN layer is proposed to estimate the abundances per-pixel by leveraging the multinomial mixture kernel. This kernel particularly provides robustness for the assumption of spectral variability since the distribution-based assumptions can be more convenient to handle to these adverse conditions as explained in the literature. Moreover, an extra NN model is utilized to capture the uncertainty term in the formulation of the spectral variability. Third, the  Wasserstein GAN is exploited in the parameter optimization which leads to more accurate and stable performance than the KL-divergence and the ME. Lastly, the model is formulated as a full pipeline based on a NN method that can be learned by backpropagation and a stochastic gradient solver.

The experimental results validate that the proposed method outperforms well-known hyperspectral unmixing baselines by a large margin. In particular, the methods obtains state-of-the-art performances on the real datasets where severe spectral variability can be observed.

\noindent
\textbf{Complexity:} As analyzed in~\cite{ozkan2017endnet, bottou2010large}, a NN model with the stochastic gradient algorithm significantly decreases the complexity and memory requirement needed by the model. More precisely, these parameters are independent from the data size due to the batch-based learning. Hence, the method is more applicable for large-scale data.

\section*{Acknowledgment}

The authors gratefully acknowledge the support of NVIDIA Corporation with the donation of GPUs used for this research.

\bibliographystyle{IEEEtran}
\bibliography{template}

%




\end{document}